\documentclass[lettersize,journal]{IEEEtran}
\usepackage{amsmath,amsfonts}
\usepackage{algorithmic}
\usepackage{algorithm}
\usepackage{array}
\usepackage[caption=false,font=normalsize,labelfont=sf,textfont=sf]{subfig}
\usepackage{textcomp}
\usepackage{stfloats}
\usepackage{url}
\usepackage{verbatim}
\usepackage{graphicx}
\usepackage{cite}
\usepackage{multirow}

\usepackage{cite}
\usepackage{pifont}
\newcommand{\cmark}{\ding{51}}  
\newcommand{\xmark}{\ding{55}}  

\usepackage{rotating}
\usepackage{booktabs, xcolor}

\usepackage{tikz}

\usepackage[utf8]{inputenc}
\usepackage[colorlinks=true, 
            urlcolor=mypink,      
            citecolor=blue, 
            linkcolor=blue]{hyperref}

\definecolor{mypink}{RGB}{220, 20, 120}

\begin{document}

\title{EM$^{2}$LDL: A Multilingual Speech Corpus for Mixed Emotion Recognition through Label Distribution Learning}

\author{Xingfeng Li, Xiaohan Shi, Junjie Li, Yongwei Li, Masashi Unoki,~\IEEEmembership{Member, IEEE}, Tomoki Toda,~\IEEEmembership{Senior Member,~IEEE}, Masato Akagi,~\IEEEmembership{Life Member, IEEE}
\thanks{This paper was produced by the IEEE Publication Technology Group. They are in Piscataway, NJ.}
\thanks{This work were supported by the JSPS Grant-in-Aid for Scientific Research (A) (25H01139) and the Beijing Natural Science Foundation (L257021).}
\thanks{Xingfeng Li and Junjie Li are with the Faculty of Data Science, City University of Macau, Macau 999078, China(e-mail: xfli@cityu.edu.mo; D24091110921@cityu.edu.mo)}

\thanks{Xiaohan Shi is with the Graduate School of Informatics, Nagoya University, Nagoya 464-8601, Japan(e-mail: xiaohan.shi@g.sp.m.is.nagoya-u.ac.jp)}

\thanks{Yongwei Li is with the Institute of Psychology, Chinese Academy of Sciences, Beijing 100101, China (e-mail: liyw@psych.ac.cn)} 

\thanks{Tomoki Toda is with the Information Technology Center, Nagoya University, Nagoya 464-8601, Japan(e-mail: tomoki@icts.nagoya-u.ac.jp)}

\thanks{Masashi Unoki and Masato Akagi are with the Graduate School of Advanced Science and Technology, Japan Advanced Institute of Science and Technology, Nomi 923-1292, Japan(e-mail: unoki@jaist.ac.jp; akagi@jaist.ac.jp)}

\thanks{Manuscript received April 19, 2021; revised August 16, 2021.}
}

\markboth{Journal of \LaTeX\ Class Files,~Vol.~14, No.~8, August~2021}%
{Shell \MakeLowercase{\textit{et al.}}: A Sample Article Using IEEEtran.cls for IEEE Journals}

\IEEEpubid{0000--0000/00\$00.00~\copyright~2021 IEEE}

\maketitle

\begin{abstract}
This study introduces EM$^2$LDL, a novel multilingual speech corpus designed to advance mixed emotion recognition through label distribution learning. Addressing the limitations of predominantly monolingual and single-label emotion corpora \textcolor{black}{that restrict linguistic diversity, are unable to model mixed emotions, and lack ecological validity}, EM$^2$LDL comprises expressive utterances in English, Mandarin, and Cantonese, capturing the intra-utterance code-switching prevalent in multilingual regions like Hong Kong and Macao. The corpus integrates spontaneous emotional expressions from online platforms, annotated with fine-grained emotion distributions across 32 categories. Experimental baselines using self-supervised learning models demonstrate robust performance in speaker-independent gender-, age-, and personality-based evaluations, with HuBERT-large-EN achieving optimal results. By incorporating linguistic diversity and ecological validity, EM$^2$LDL enables the exploration of complex emotional dynamics in multilingual settings. This work provides a versatile testbed for developing adaptive, empathetic systems for applications in affective computing, including mental health monitoring and cross-cultural communication. The dataset, annotations, and baseline codes are publicly available at \url{https://github.com/xingfengli/EM2LDL}.
\end{abstract}


\section{Introduction}
\IEEEPARstart{S}peech emotion recognition (SER) is a cornerstone of affective computing, enabling machines to interpret human emotional states through vocal cues \cite{wu2010emotion,alisamir2021evolution}. This field has garnered significant attention due to its potential to enhance human-computer interaction, support mental health monitoring, and facilitate cross-cultural communication \cite{pantic2003toward,jeon2024effects}. With applications spanning psychology, education, and customer service, SER plays a vital role in developing empathetic and responsive technologies. Significant progress has been made in SER through the application of machine learning and representation learning methods, including supervised, self-supervised, and semi-supervised techniques \cite{deng2017semisupervised,latif2021survey}, thereby establishing SER as a key driver of real-time sentiment analysis and automated mental health assessments \cite{chatterjee2021real}.

\IEEEpubidadjcol
Despite these advancements, two critical challenges hinder the progress of SER toward broader, real-world applicability. First, most SER research has focused on monolingual contexts, predominantly targeting English, Mandarin, or other widely spoken individual languages \cite{mohamed2022self,akccay2020speech}. However, over half of the world’s population is bilingual or multilingual, and approximately 50–60\% of countries recognize two or more official languages in governance, education, or public services, where speakers frequently code-switch or use multiple languages within a single interaction \cite{jayanath2021comparative,grosjean2024bilinguals,gurevich2025dataset}. In countries such as India, Switzerland, and Singapore, or regions like Hong Kong and Macao, speakers routinely switch between languages within a single interaction, a phenomenon known as code-switching or intra-utterance multilingualism \cite{goh2016issue, lising2020code,liu2024enhancing}. Emotional expression in such environments is further complicated by cultural and linguistic variability \cite{fatima2024language,mavrou2020emotionality}. Despite the widespread prevalence of multilingual societies, relatively few SER studies have addressed emotion recognition across languages \cite{gerczuk2021emonet,gao2023adversarial}, creating a substantial gap in both datasets and algorithms.

\textcolor{black}{Second, conventional SER approaches typically rely on single-label emotion classification assuming that an utterance conveys a single dominant emotion \cite{latif2021survey,akccay2020speech}}. In reality, human emotions are often complex and mixed, with utterances frequently expressing multiple emotions simultaneously \cite{russell2017mixed,zhou2022speech,schuller2023acm}. \textcolor{black}{For example, naturalistic vocal expressions may even combine opposite valence paired ones, such as happiness and sadness, as seen in bittersweet farewells, or fear and hope, as in anxious anticipation \cite{israelsson2023blended,bee2013consumer}.} Studies have shown that approximately 41\% of naturalistic recordings reflect mixed emotional states, highlighting the prevalence of mixed emotions \cite{juslin2021spontaneous}. Complementary research in nonverbal behavior has also documented the co-activation of happiness and fear \cite{andrade2007consumption}, amusement and disgust \cite{hemenover2007s}, and other mixed emotional states \cite{larsen2011further}. Such findings highlight the limitations of conventional single-label SER systems, which are inherently ill-equipped to model the full spectrum of human affect \cite{schuller2023acm,kang2020active}.

To address these gaps, we introduce EM$^{2}$LDL, a novel \textbf{E}motion-labeled \textbf{M}ultilingual \textbf{M}ixed-emotion corpus that leverages \textbf{L}abel \textbf{D}istribution \textbf{L}earning for emotion modeling. EM$^{2}$LDL captures fine-grained emotional nuance through probabilistic distributions over multiple emotion labels per utterance, thus reflecting the co-existence and intensity of mixed emotional states. Furthermore, this corpus incorporates intra-utterance multilingualism, where multiple languages are spoken within a single utterance. It thereby enables SER in linguistically dynamic contexts. The EM$^{2}$LDL comprises three languages: English, Mandarin, and Cantonese, with intra-utterance code-switching settings, and is annotated with emotion label distributions by human raters spanning diverse backgrounds.

In addition to the corpus itself, we propose an evaluation framework that includes baseline systems, multilingual validation protocols, and reproducibility-friendly experimental designs. EM$^{2}$LDL is thus intended not only as a resource for benchmarking multilingual mixed emotion recognition but also as a step toward building effective technologies that are inclusive, culturally adaptable, and ecologically valid. This work has far-reaching implications for multilingual societies, where emotion-aware systems must navigate linguistic diversity and emotional complexity simultaneously.

Section II of this paper reviews related work on multilingual and mixed SER. In Section III, we introduce the construction, annotation, and statistical analysis of the EM$^{2}$LDL corpus. Section IV presents the experimental setup and baseline evaluations, and Section V discusses the implications of our findings and outlines directions for future research. We conclude in Section VI with a brief summary.

\section{Related Work}
\subsection{Existing Emotion Speech Corpora}
A diverse range of SER corpora have been developed over the past two decades, serving as the foundation for progress in affective computing. These corpora differ in terms of language coverage, emotional categories, elicitation strategies, recording environments, and annotation protocols \cite{latif2021survey,mohamed2022self,gerczuk2021emonet}. Pioneering resources such as IEMOCAP \cite{busso2008iemocap}, EmoDB \cite{burkhardt2005database}, CASIA \cite{zhang2008design}, and MSP-Podcast \cite{martinez2020msp} have significantly contributed to the field by providing high-quality, annotated emotional speech under controlled or semi-naturalistic conditions. However, three major limitations are consistently observed across most existing corpora.
\begin{itemize}
\item \textcolor{black}{First, they are predominantly monolingual, thus limiting their capacity to support multilingual SER or to reflect the cross-linguistic diversity of emotional expression.}
\item \textcolor{black}{Second, most of these datasets utilize single-label categorical annotations, assuming that each utterance conveys only one dominant emotion. This assumption fails to capture the complex and overlapping nature of real-life affective expressions.}
\item \textcolor{black}{Third, many corpora rely on acted speech samples which, despite their consistency, tend to lack the spontaneity and ecological validity of naturalistic interactions.}
\end{itemize}

These constraints reduce the generalizability of SER models in realistic, multicultural, and emotionally nuanced scenarios. Several recent corpora are trying to address one or more of these issues. For example, SEWA \cite{kossaifi2019sewa} and CMU-MOSEAS \cite{zadeh2020cmu} incorporate multilingual and multimodal content, and MSP-Podcast \cite{martinez2020msp} provides continuous emotional labels. However, none of these resources systematically integrate the three critical dimensions of multilingualism, spontaneous speech, and mixed-emotion annotation into a unified framework.

Table~\ref{tab:emotion_corpora} presents a comparative overview of representative emotion speech corpora, highlighting the lack of resources that jointly support all three dimensions. This deficiency underscores the need for SER corpora that more accurately reflect the sociolinguistic and emotional complexities of real-world communication.
\begin{table*}[!t]
\centering
\caption{Comparative Summary of EM$^{2}$LDL and Representative Emotion Speech Corpora}
\label{tab:emotion_corpora}
\small
\resizebox{\textwidth}{!}{
\begin{tabular}{l c c c l c l l}
\toprule
\textbf{Corpus} & \textbf{Language(s)} & \textbf{Multilingual} & \textbf{Code-Switching} & \textbf{Expression Type} & \textbf{Mixed Emotion} & \textbf{Label Type} & \textbf{Notes} \\
\midrule
IEMOCAP        & En                   & \xmark & \xmark                     & Scripted-Semi-Natural & \xmark & Category, VAD           & Dyadic, multimodal       \\
EmoDB          & De                   & \xmark & \xmark                     & Acted                 & \xmark & Category                & Studio-quality recordings \\
CASIA          & Zh                   & \xmark & \xmark                     & Acted                 & \xmark & Category                & Balanced Mandarin corpus \\
MSP-Podcast    & En                   & \xmark & \xmark                     & Naturalistic          & Partial & Category, VAD         & Podcast-based spontaneous speech \\
MELD           & En                   & \xmark & \xmark                     & Naturalistic          & \xmark & Category                & Multi-party TV dialogues \\
SEWA           & Zh, En, De, El, Hu, Sr & \cmark & \xmark                  & Naturalistic          & \xmark & VAL                     & Multilingual video interviews \\
CMU-MOSEAS     & Fr, Es, Pt, De       & \cmark & \xmark                     & Naturalistic          & \xmark & Category                & Topic-diverse public speeches \\
EmoV-DB        & En, Fr               & \cmark & \xmark                     & Acted                 & \xmark & Category                & Designed for TTS training \\
\textbf{EM$^{2}$LDL} & \textbf{En, Zh, Yue} & \textbf{\cmark} & \textbf{\cmark~(Intra-Utterance)} & \textbf{Naturalistic} & \cmark & \textbf{Label Distribution} & \textbf{Code-switching \& mixed LDL} \\
\bottomrule
\end{tabular}
}

\vspace{0.5em}
\begin{minipage}{\textwidth}
\footnotesize
\textit{Language abbreviations}: En: English, Zh: Mandarin, Yue: Cantonese, De: German, Es: Spanish, Pt: Portuguese, Fr: French, El: Greek, Hu: Hungarian, Sr: Serbian.  
\textit{Symbol legend}: \cmark = Supported; \xmark = Not supported; Partial = Partially supported.  
\textit{Label types}: VAD = Valence-Arousal-Dominance; VAL = Valence-Arousal-Liking; Category = Discrete emotion categories.  
\textit{Code-switching}: Alternation of languages within a single utterance.
\end{minipage}
\end{table*}

\subsection{Multilingual Speech Emotion Recognition}
As speech-based AI technologies become increasingly global, multilingual SER has emerged as a critical area of investigation \cite{wang2023multilingual,sharma2022multi}. Prior studies have implemented cross-lingual transfer learning, multilingual fine-tuning, and domain adaptation techniques to improve generalization across languages \cite{gao2023adversarial,latif2022self,feng2024trust}. For example, models trained on English datasets have been adapted to recognize emotions in Mandarin or German through adversarial domain alignment or multilingual embeddings \cite{upadhyay2025phonetically,wang2025learning}. While these approaches have demonstrated the feasibility of language-agnostic SER, they often rely on monolingual datasets from distinct domains.

A major limitation of current multilingual SER research is its treatment of language boundaries as static and mutually exclusive \cite{sharma2022multi,li2019improving,goncalves2024bridging}. In many multilingual communities—including those in India, Singapore, Hong Kong, and Macao—speakers frequently engage in intra-utterance code-switching, fluidly mixing languages within the same sentence \cite{sitaram2019survey,smolak2020code}. This phenomenon reflects natural communicative behavior but is rarely accounted for in SER research due to a lack of representative data. Most existing studies assume clear language separation and do not consider the linguistic dynamics of code-switching, thus overlooking an important dimension of real-world multilingualism.

Furthermore, emotional expression is known to be culturally and linguistically shaped \cite{sadiqzade2025linguistic,ibrakhim2004universal}, and as such, emotional expressions may manifest differently across languages and are interpreted within specific sociolinguistic contexts. Without datasets that capture intra-utterance multilingualism and cultural variability, current SER systems risk misinterpreting or oversimplifying emotional signals in diverse populations.

\begin{table*}[!t]
\centering
\caption{Comparison of Emotion Modeling Techniques for Multilingual Mixed Emotion Recognition}
\label{tab:mixed_methods_summary}
\small
\resizebox{\textwidth}{!}{
\begin{tabular}{l l c c l l}
\toprule
\textbf{Method} & \textbf{Label Type} & \textbf{Mixed Emotion} & \textbf{Multilingual} & \textbf{Strengths} & \textbf{Limitations} \\
\midrule
Single-label Classification   & One-hot categorical       & \xmark   & \cmark   & Simple implementation, widely adopted & Cannot model co-occurrence or ambiguity \\
Multi-label Classification    & Binary vector (0/1)       & \cmark   & Partial  & Supports co-occurrence modeling        & Ignores emotion intensity and correlation \\
Valence-Arousal (VA)          & 2D continuous values       & \cmark   & Partial  & Effective for subtle affect variations & Lacks discrete interpretability, cannot capture opposing values \\
Soft Labeling                 & Real-valued vector         & \cmark   & Partial  & Captures rater uncertainty             & No standard label distribution framework \\
Label Distribution Learning (LDL) & Probability distribution & \cmark   & Partial  & Fine-grained, interpretable, probabilistic & Requires specialized data and annotations \\
\bottomrule
\end{tabular}
}

\vspace{0.5em}
\begin{minipage}{\textwidth}
\footnotesize
\textit{Symbol legend}: \cmark = Supported; \xmark = Not supported; Partial = Partially supported.  
\textit{Multilingual}: Indicates whether the modeling technique is commonly applied or well supported in multilingual SER settings.
\end{minipage}
\end{table*}

\subsection{Mixed Emotion Modeling in Speech}
In contrast to the conventional view that emotions are discrete and mutually exclusive \cite{ekman1994nature}, psychological and neurocognitive studies increasingly support the perspective that emotional states are often mixed \cite{watson2017emotion,zhao2022unpacking,kreibig2013psychophysiology}. That is, individuals frequently experience multiple affective dimensions simultaneously. For example, sarcastic speech may convey both amusement and irritation \cite{kader2022computational}, while public performance can evoke both pride and anxiety \cite{cohen2018positive}. Empirical work by Cowen and Keltner \cite{cowen2017self} revealed that emotional responses frequently span multiple categories, even when triggered by a single stimulus. These findings challenge the adequacy of conventional single-label classification frameworks in capturing the richness and complexity of human affect.

To address this limitation, a range of computational approaches have been proposed to model mixed emotions. These include multi-label classification \cite{li2022multi,deng2020multi}, which allows assigning multiple categorical tags per utterance; dimensional models such as VA space \cite{nicolaou2011continuous,valenza2011role}, which represent affective states as points in a continuous space; and soft labeling schemes that encode rater uncertainty as real-valued distributions \cite{mao2021enhancing,fard2024affectnet+}. More recently, label distribution learning (LDL) has emerged as a promising framework that jointly models both the co-occurrence and intensity of emotional components using probability distributions \cite{prabhu2023end,le2023uncertainty}. Table~\ref{tab:mixed_methods_summary} summarizes and compares these representative modeling strategies in terms of label encoding, support for mixed emotion modeling, multilingual applicability, and key trade-offs.

Despite the conceptual advances, several limitations remain. Many of these methods have been developed and evaluated using monolingual datasets, which limits their generalizability to multilingual or cross-cultural contexts \cite{wen2023learning,kumawat2025extending}. Moreover, there is significant fragmentation in annotation practices. Some studies rely on rater agreement thresholds to define dominant emotion labels, while others permit open-ended emotion tagging or aggregate ratings probabilistically \cite{juslin2005vocal,wu2014survey}. The lack of standardized, reproducible annotation protocols undermines the comparability and scalability of mixed emotion recognition systems. These challenges highlight the urgent need for high-quality speech corpora that offer consistent, theoretically grounded mixed emotion annotations and support multilingual emotion modeling under real-world conditions.

\subsection{Label Distribution Learning for Emotion Recognition}
LDL offers a principled framework for modeling fuzzy, ambiguous, or overlapping affective phenomena \cite{le2023uncertainty,maestro2024towards,khelifa2025label}. Unlike conventional single-label or multi-label classification, LDL assigns a probability distribution over all possible emotion labels, thereby quantifying the degree to which each emotion is present in a given sample. This approach is particularly well-suited for SER tasks involving subtle or compound emotions, where no single label is sufficient to describe the affective state.

These days, LDL has been successfully applied in visual emotion recognition tasks, such as facial expression analysis using the AffectNet and RAF-DB datasets \cite{fard2024affectnet+,shin2024noisy}. In the speech domain, recent works have begun exploring LDL for modeling soft valence-arousal values or estimating emotion intensity distributions from crowd-sourced annotations \cite{schuller2023acm,le2023uncertainty,kim2025effect}. These studies demonstrate that LDL can improve prediction performance and better reflect human perception.

However, the use of LDL in multilingual and mixed-emotion speech recognition remains scarce. Most existing applications are confined to single-language datasets with controlled lab settings. There is currently no corpus that enables the joint exploration of LDL-based modeling, multilingual emotional expression, and intra-utterance language mixing. This lack of integrated resources limits the broader adoption of LDL for affective computing in global and heterogeneous user contexts.

\subsection{Motivating Insights and EM$^{2}$LDL Contribution}
In line with these findings, three key gaps persist across the existing literature on SER: (1) current emotion corpora are predominantly monolingual, acted, and single-labeled; (2) multilingual SER lacks corpora that reflect intra-utterance code-switching and sociolinguistic variability; and (3) although LDL shows promise for emotion modeling, it remains underutilized in multilingual mixed-affect speech.

To address these challenges, the EM$^{2}$LDL corpus makes three primary contributions to the field of affective computing:

\begin{itemize}
\item \textcolor{black}{\textbf{Multilingual and Code-Switching Support}: It is the first publicly available corpus to include intra-utterance code-switching among English, Mandarin, and Cantonese, reflecting the linguistic reality of multilingual societies.}
\item \textcolor{black}{\textbf{Mixed-Emotion Modeling via LDL}: It provides probabilistic emotion distributions for each utterance, enabling the modeling of overlapping and ambiguous emotional states.}
\item \textcolor{black}{\textbf{Ecological Validity and Demographic Diversity}: It comprises spontaneous speech from online platforms with balanced speaker demographics and personality annotations, thereby supporting research in personalized and culturally aware SER.}
\end{itemize}

The proposed EM$^{2}$LDL not only contributes a unique resource to the field but also facilitates the development of more generalizable and interpretable emotion recognition models. In doing so, it opens up new research directions at the intersection of multilingualism and emotional complexity, thereby addressing pressing challenges in both data construction and algorithmic modeling within affective computing.
\begin{figure}[t]  
\centering
\includegraphics[width=\columnwidth]{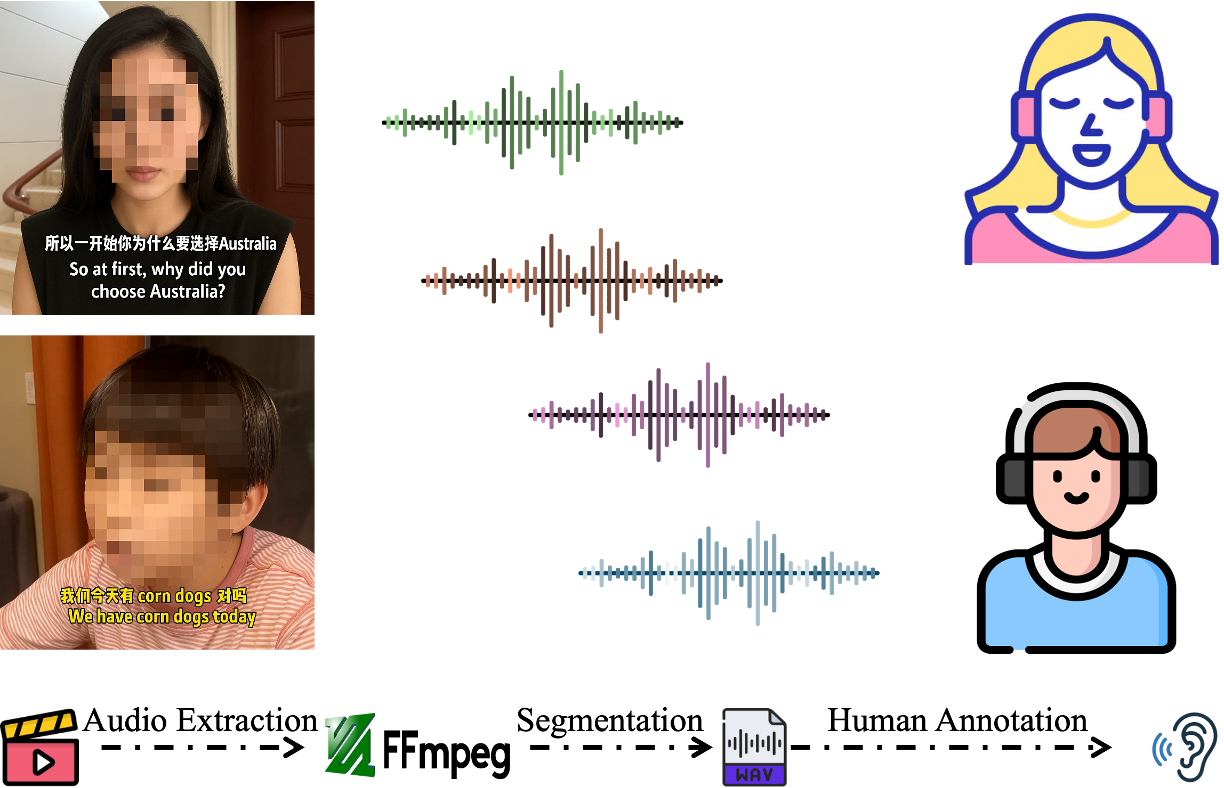}  
\caption{Overview of the multilingual data acquisition and annotation pipeline. Video segments containing multilingual and code-switched emotional speech are collected from online platforms (left). Audio tracks are extracted and segmented into utterances (center). The resulting speech samples are then annotated by human raters using a label distribution format (right).}
\label{fig:data_pipeline}
\end{figure}

\section{EM$^{2}$LDL Corpus}
\subsection{Design Motivation and Objectives}
The EM$^{2}$LDL corpus is conceived as a targeted response to three intertwined challenges in current SER research: linguistic homogeneity, oversimplified emotional representation, and limited support for probabilistic modeling frameworks. Building on our review of these limitations from a literature perspective in Section II, this section formalizes them as design drivers that guide the corpus construction.

More specifically, EM$^{2}$LDL aims to achieve the following design objectives:

\begin{itemize}
\item \textcolor{black}{\textbf{Linguistic Diversity}: Include three typologically distinct languages by considering English (non-tonal), Mandarin (tonal), and Cantonese (tonal with high variability) to support multilingual and cross-linguistic SER evaluation.}
\item \textcolor{black}{\textbf{Code-Switching Realism}: Incorporate naturally occurring intra-utterance code-switching scenarios to reflect multilingual and diglossic communication patterns commonly found in globalized contexts.}
\item \textcolor{black}{\textbf{Emotion Distribution Fidelity}: Move beyond categorical or binary labels by adopting an LDL framework, where each utterance is annotated with a fine-grained emotional probability distribution derived from multiple human raters.}
\end{itemize}

By aligning the EM$^{2}$LDL corpus construction with both psychological realism and computational feasibility, these design choices are not merely compensatory: they enable novel research directions in affective computing, particularly in modeling emotional ambiguity, cross-lingual generalization, and probabilistic inference. 

\begin{figure}[t]  
\centering
\includegraphics[width=\columnwidth]{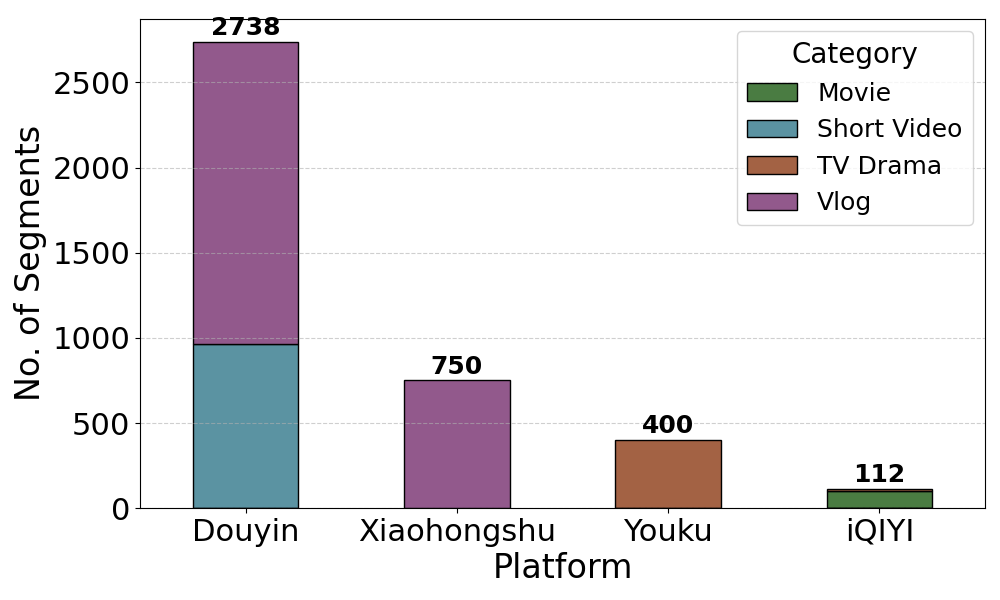}  
\caption{Distribution of speech segments across online platforms and content categories.}
\label{fig:platform_dist}
\end{figure}

\subsection{Data Sources and Collection Strategy}
The EM\textsuperscript{2}LDL corpus is designed to capture spontaneous emotional expressions in real-world multilingual contexts, distinguishing itself from conventional laboratory-recorded corpora that typically rely on scripted or acted speech. To achieve ecological validity and reflect linguistic diversity, we sourced audio data from four prominent social and video-sharing platforms in China: Douyin, Xiaohongshu, Youku, and iQIYI. These platforms were chosen for their extensive user engagement, rich multimodal content, and prevalence of naturalistic emotional interactions, providing a robust foundation for the corpus.

The data collection process commenced with the extraction of over 4,000 utterances in English, Mandarin, and Cantonese, selected on the basis of stringent linguistic and affective criteria to encompass a wide spectrum of emotional tones and communication contexts. Particular attention was given to preserving intra-utterance code-switching, a common linguistic phenomenon among bilingual or trilingual speakers, especially in regions such as Hong Kong and Macao \cite{li2020tale,choi2024english,mak2020multilingualism}. This approach ensures the corpus mirrors the sociolinguistic realities of multilingual communities.

To visually encapsulate the initial phase of this process, Fig.~\ref{fig:data_pipeline} illustrates the workflow, featuring representative video frames from online platforms and their corresponding audio waveforms extracted using \textcolor{black}{FFmpeg\footnotemark[1]}. This technique preserves the natural variability of speech patterns across languages, laying the groundwork for subsequent analysis. The right panel of Fig.~\ref{fig:data_pipeline} previews the transition to the annotation phase, where human raters were engaged to assign emotion label distributions, thereby capturing the complexity of mixed emotional states. \footnotetext[1]{\url{https://www.ffmpeg.org/download.html\#get-sources}}

Subsequently, all audio segments underwent rigorous manual screening to ensure emotional relevance, acoustic clarity, and linguistic integrity. Utterances were segmented at the speaker-turn level to maintain contextual independence and mitigate dialogic interference. To standardize the dataset for downstream modeling, waveforms were resampled to a \textcolor{black}{16-kHz} single-channel 16-bit PCM format, ensuring compatibility and consistency across analyses. Overall, Fig.~\ref{fig:platform_dist} provides a detailed breakdown highlighting the varying contributions of each source to the EM\textsuperscript{2}LDL corpus.

In contrast to existing emotion corpora that often depend on controlled role-play or induced emotions, the EM\textsuperscript{2}LDL corpus prioritizes ecological authenticity, sociolinguistic realism, and linguistic heterogeneity. This design supports a diverse array of modeling paradigms, including multilingual SER, code-switched speech processing, and mixed-emotion analysis through LDL, positioning it as a valuable resource for future affective computing research.
\begin{figure}[t]  
\centering
\includegraphics[width=\columnwidth]{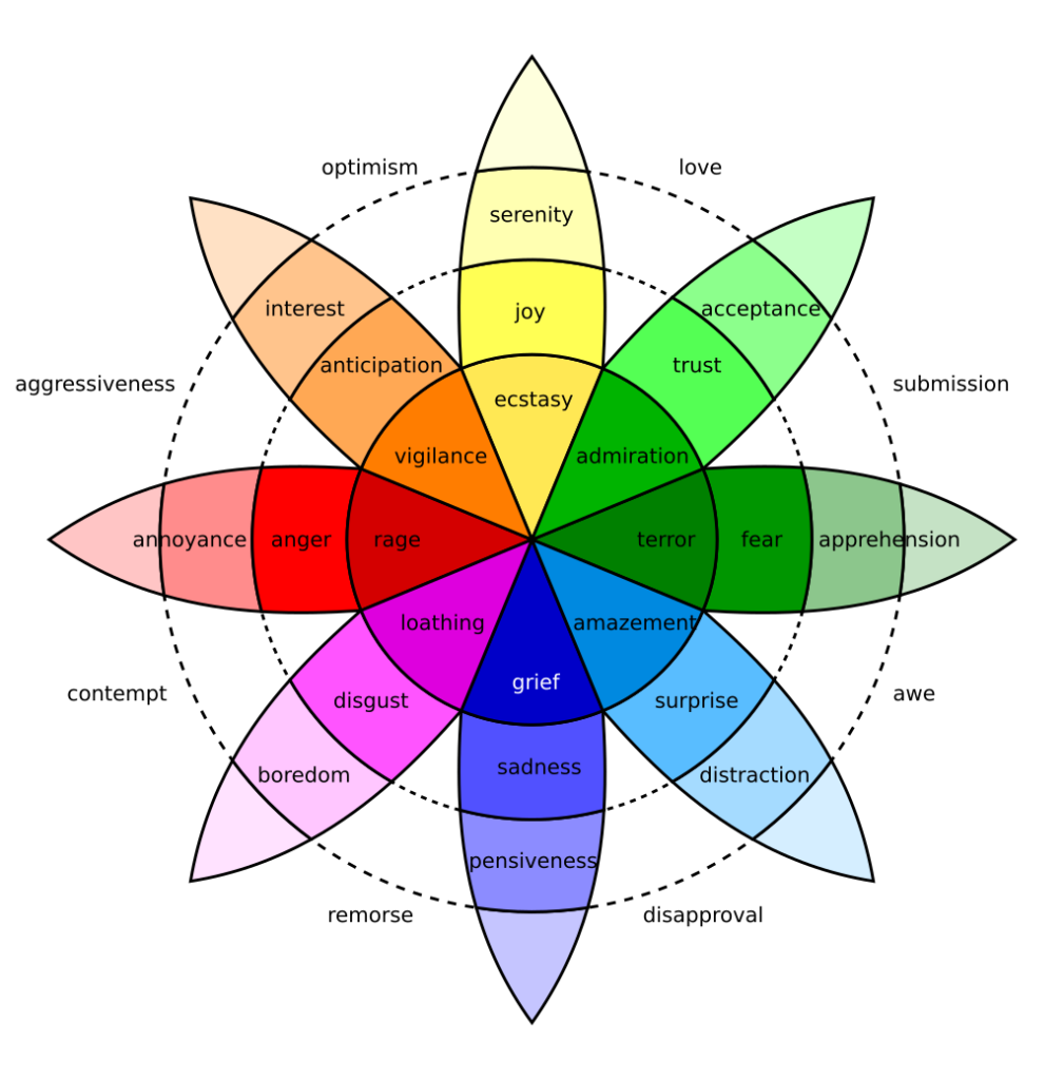}  
\caption{Illustration of Plutchik’s emotion wheel used as the basis for mixed emotion annotation.}
\label{fig:Plutchiksemotion}
\end{figure}

\begin{table}[t]
\centering
\caption{Demographic and Linguistic Background of Human Raters}
\label{tab:rater_info}
    \resizebox{\columnwidth}{!}{
    \begin{tabular}{ccccc}
        \toprule
        \textbf{Human Raters} & \textbf{Gender} & \textbf{MBTI} & \textbf{Age} & \textbf{Languages} \\
        \midrule
        1 & Female & ISFP & 21 & Mandarin, English, Cantonese \\
        2 & Male & ISFJ & 22 & Mandarin, English, Cantonese \\
        3 & Male & ISTP & 22 & Mandarin, English \\
        4 & Male & ENTP & 22 & Mandarin, English \\
        5 & Male & INFP & 22 & Mandarin, English \\
        6 & Male & INFP & 23 & Mandarin, English \\
        7 & Female & ENFJ & 23 & Mandarin, English \\
        8 & Male & ISFP & 22 & Mandarin, English \\
        9 & Male & INTJ & 23 & Mandarin, English \\
        10 & Male & ISTJ & 24 & Mandarin, English \\
        11 & Female & INFJ & 21 & Mandarin, English \\
        12 & Male & ENFJ & 22 & Mandarin, English \\
        13 & Male & ESTP & 25 & Mandarin, English, Cantonese \\
        14 & Female & ISTJ & 19 & Mandarin, English \\
        15 & Female & INTP & 21 & Mandarin, English \\
        16 & Male & ENFJ & 22 & Mandarin, English, Cantonese \\
        17 & Male & ISTJ & 22 & Mandarin, English, Cantonese \\
        18 & Female & ENFP & 22 & Mandarin, English, Cantonese \\
        19 & Female & INFP & 22 & Mandarin, English, Cantonese \\
        20 & Male & ISFP & 22 & Mandarin, English, Cantonese \\
        \bottomrule
    \end{tabular}
    }
\end{table}

\subsection{Annotation Protocol and Label Distribution Design}
To support the development of mixed-emotion recognition models under an LDL paradigm, all utterances in the $\mathrm{EM}^{2}\mathrm{LDL}$ corpus were annotated by 20 independent human raters using a structured perceptual protocol. The annotation process was designed to capture co-occurrence emotions with varying intensity levels while preserving interrater diversity, ensuring a robust representation of emotional complexity in multilingual contexts.

\subsubsection{Annotation Protocol}  
Emotion perception was guided by Plutchik’s Emotion Wheel, a comprehensive framework that categorizes 32 distinct emotions, as illustrated in Fig.~\ref{fig:Plutchiksemotion}. This model was selected for its ability to represent a wide range of emotions and their interrelationships, making it particularly suitable for capturing the nuanced, mixed emotional states prevalent in code-switched and multilingual speech \cite{wang2020emo2vec,molina2019improving}. By leveraging Plutchik’s framework, our annotation protocol ensures that subtle emotional variations, such as combinations of joy and trust or anger and fear, are systematically identified and quantified, providing a psychologically grounded basis for modeling complex emotional expressions. \textcolor{black}{Raters were instructed to listen to each stimulus at the sentence level individually and indicate all perceived emotions by marking a binary value (1 or 0) in a standardized spreadsheet template.} The auditory stimuli and annotation templates were organized sequentially to ensure labeling consistency. All participants were fluent in at least two of the corpus languages (Mandarin, Cantonese, or English), and their demographic backgrounds spanned gender, age, and linguistic profiles (as summarized in Table~\ref{tab:rater_info}), thereby enhancing the representativeness of emotional perception.

\subsubsection{Label Distribution Computation}  
To transform the multi-rater binary annotations into soft emotion distributions, we adopted a frequency-based normalization strategy following \cite{schuller2023acm,oatley2014communicative,mcgraw2010benign}, namely, inferring the concurrent experience and intensity of mixed emotions by quantifying the frequency with which individuals experience each measured emotion. 

Formally, for each utterance $n$ and all emotion classes $C$, we constructed an annotation matrix $\mathrm{M}$ of size $R \times C$:

\begin{equation}
\mathrm{M} = 
\begin{bmatrix}
\epsilon_{1,1}^n & \cdots & \epsilon_{1,C}^n \\
\vdots & \ddots & \vdots \\
\textcolor{black}{\epsilon_{R,1}^n} & \cdots & \textcolor{black}{\epsilon_{R,C}^n}
\end{bmatrix}, \quad \epsilon_{r,c}^n \in \{0,1\},
\end{equation}
where $C$ = 32 is the total number of emotional states and  \textcolor{black}{$R$ = 20 refers to human annotators}. Each $\epsilon_{r,c}^n$ reflects whether rater $r$ perceived emotion $c$ for utterance $n$. The final label distribution for utterance $n$ over a certain emotion class $c$ is computed as

\begin{equation}
\rho_{c}^n = \frac{\sum_{r=1}^{20} \epsilon_{r,c}^n}{\sum_{r=1}^{20} \sum_{c} \epsilon_{r,c}^n}.
\end{equation}

\begin{figure}[t]  
\centering
\includegraphics[width=\columnwidth]{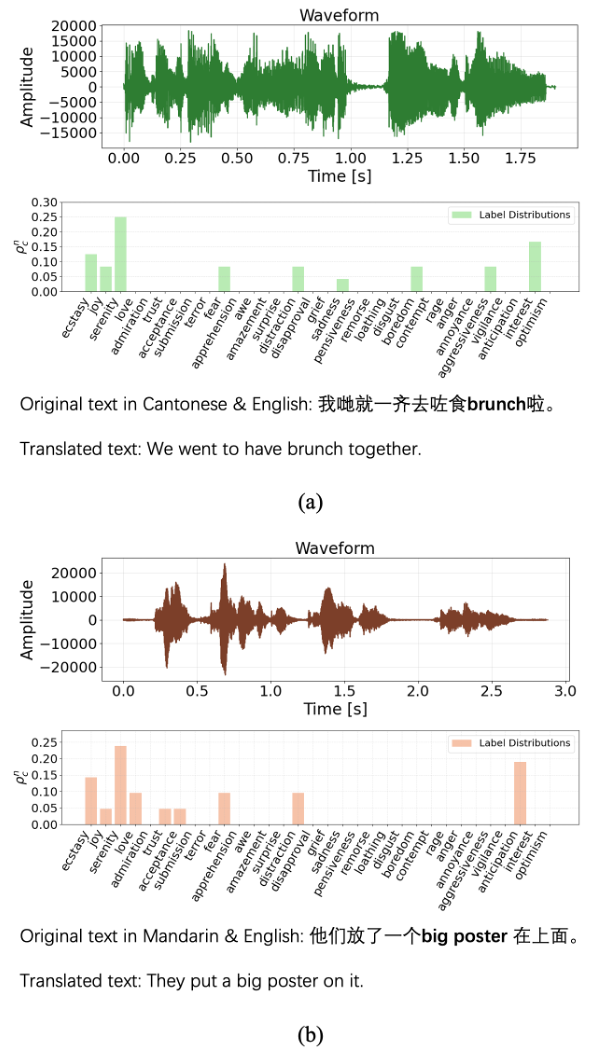}  
\caption{\textcolor{black}{Label distribution examples computed from human ratings for two speech samples.}}
\label{fig:label_distribution}
\end{figure}

Figure~\ref{fig:label_distribution} shows an example of the $\rho_{c}^n$ relative to two utterance stimuli in the $\mathrm{EM}^{2}\mathrm{LDL}$ corpus. The X-axis represents the 32 kinds of emotions and the Y-axis represents the intensities of each kind of emotion. Note that $\rho_{c}^n$ denotes the proportion that $c$ accounts for in a full emotion distribution of the $n^{th}$ utterance stimulus. This formulation yields a probability distribution over the emotion space that reflects both the co-occurrence and relative intensity of perceived emotions. Unlike probabilistic hard labeling, which assumes a single dominant class, this approach enables the modeling of mixed emotions while preserving perceptual diversity across annotators.

\subsection{Corpus Statistics and Analysis}
\subsubsection{Audio Characteristics}
The $\mathrm{EM}^{2}\mathrm{LDL}$ corpus contains a total of 3,998 audio utterances, amounting to 14,540.08 seconds of speech (approximately 4.04 hours). The average duration per utterance is 3.64 seconds, reflecting the concise yet emotionally expressive nature of the collected segments. Note that although 4,000 utterances were originally collected, two audio files were excluded due to corruption detected during the annotation process. The corpus captures intra-utterance code-switching across three language pairs: Cantonese-English (CE), Mandarin-English (ME), and Mandarin-Cantonese (MC). Table~\ref{tab:audio_stats} summarizes the distribution of utterances and their durations by code-switching type.

As shown in the table, the CE type dominates with 63.0\% of utterances and 61.9\% of total duration, enabling the study of emotional expression under the dynamic, bilingual conditions common in multilingual societies such as Hong Kong and Macao. The shorter average duration of MC utterances (approximately 0.29 h) compared to CE (2.50 h) and ME (1.25 h) suggests linguistic or contextual differences in code-switching patterns, which merit further investigation.

\subsubsection{Emotion Label Distribution}
\begin{figure}[t]  
    \centering
    \includegraphics[width=\columnwidth]{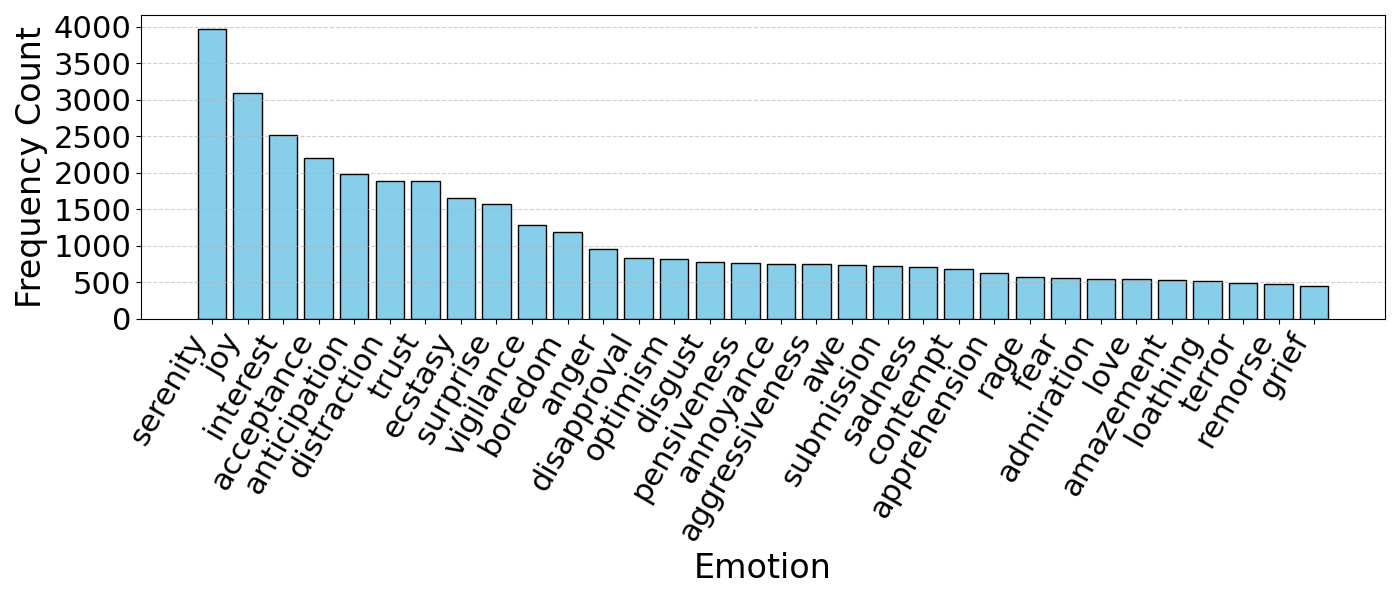}  
    \caption{Frequency distribution of each single emotional category in the whole EM$^2$LDL corpus.}
    \label{fig:emotion_freq}
\end{figure}

Each utterance in the $\mathrm{EM}^{2}\mathrm{LDL}$ corpus is annotated with a probability distribution over 32 emotion categories derived from 20-rater annotations based on Plutchik’s Emotion Wheel. On average, each utterance is associated with 9.25 emotion labels (standard deviation: 1.65), with a maximum of 16 and a minimum of 4 labels, reflecting the complexity of mixed emotional states. Figure~\ref{fig:emotion_freq} illustrates the frequency of each emotion across the corpus, with serenity (3,962 occurrences), joy (3,092), and interest (2,515) being the most prevalent, while grief (446), remorse (470), and terror (490) are less frequent.

This distribution highlights the corpus’s ability to capture a wide range of emotional expressions, with positive emotions (e.g., serenity, joy) appearing more frequently than intense negative emotions (e.g., terror, grief). The high average number of labels per utterance supports the use of LDL as it enables the modeling of co-occurrence emotions with varying intensities, thereby addressing the limitations of single-label SER systems.

\begin{table}[t]
\centering
\caption{Overview of Audio Utterances and Durations by Code-Switching Language Pairs in the EM$^2$LDL Corpus}
\label{tab:audio_stats}
\resizebox{\columnwidth}{!}{
\begin{tabular}{l c c}
\toprule
\textbf{Code-Switch Type} & \textbf{No. of Utterances} & \textbf{Total Duration (s)} \\
\midrule
Cantonese-English & 2,520 & 9,006.60 \\
Mandarin-English & 1,249 & 4,502.51 \\
Mandarin-Cantonese & 229 & 1,030.97 \\
\midrule
\textbf{Total} & 3,998 & 14,540.08 \\
\bottomrule
\end{tabular}
}
\end{table}

\subsubsection{Speaker Demographics}
\begin{figure}[t]  
    \centering
    \includegraphics[width=\columnwidth]{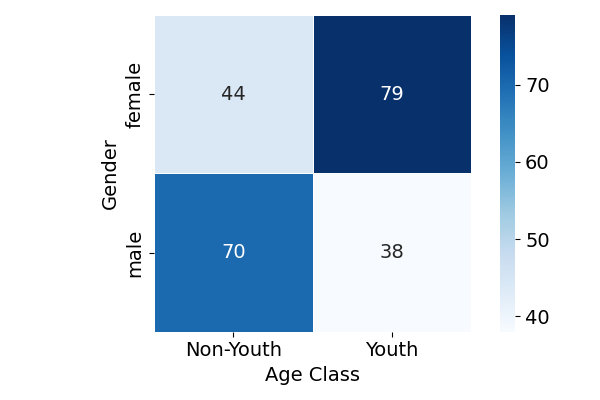}  
    \caption{Distribution of speakers across gender and age classes in the EM$^2$LDL corpus.}
    \label{fig:agS}
\end{figure}

\begin{figure}[t]  
    \centering
    \includegraphics[width=\columnwidth]{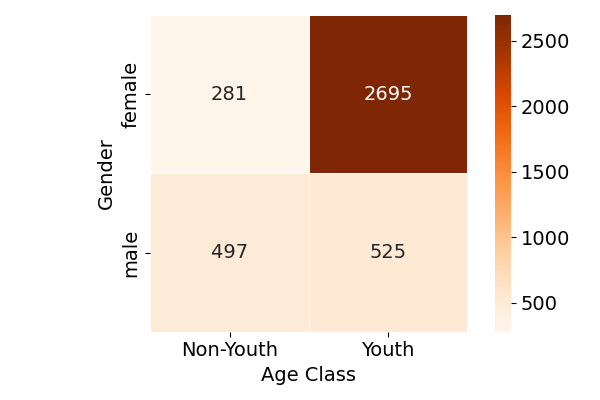}  
    \caption{Distribution of speech segments across gender and age classes in the EM$^2$LDL corpus.}
    \label{fig:agU}
\end{figure}

The $\mathrm{EM}^{2}\mathrm{LDL}$ corpus includes contributions from 231 unique speakers with a balanced gender distribution (108 male, 123 female) and \textcolor{black}{self-reported/human-rated} age classes (117 youth \textcolor{black}{15–35 years of age}, 114 non-youth). Figures~\ref{fig:agS} and~\ref{fig:agU} visualize the distribution of speakers and utterances by gender and age class. Female speakers contribute 2,976 utterances (74.4\% of the total), significantly outnumbering male speakers (1,022 utterances), while youth speakers account for 3,220 utterances (80.5\%) compared to 778 for non-youth.

The speaker with the most samples contributed 864 utterances while the least contributed just one, indicating a skewed distribution of speaker activity. This variability supports the corpus’s ecological validity, as it reflects natural participation patterns in social media contexts. The demographic diversity enhances the generalizability of SER models trained on $\mathrm{EM}^{2}\mathrm{LDL}$ across different populations.

\subsubsection{Personality Traits (MBTI)}
\textcolor{black}{Self-reported} MBTI personality data is available for a subset of speakers, where 13 (5.6\%) reported as Introversion (I), 99 (42.9\%) as Extraversion (E), and 119 (51.5\%) with unknown MBTI data. Interestingly, the utterance counts across the known personality subtypes are as follows: Extraversion-Sensing (ES) includes 773 utterances, Extraversion-Intuition (EN) includes 619 utterances, Introversion-Sensing (IS) includes 30 utterances, and Introversion-Intuition (IN) includes 43 utterances.

The prevalence of extraverted speakers aligns with the spontaneous, socially engaged nature of the source platforms, providing a unique opportunity to explore the interplay between personality traits and emotional expression in multilingual contexts. \textcolor{black}{In general, these characteristics mentioned above position $\mathrm{EM}^{2}\mathrm{LDL}$ as a strong benchmark for developing ecologically valid, multilingual, and emotionally nuanced affective computing models.}

\section{Experiment Baselines}
\subsection{Experimental Setup}
\label{sec:exp_setup}
This section establishes the evaluation protocol, acoustic representations, baseline model architecture, training procedure, and quantitative criteria used in our EM$^{2}$LDL benchmark. Our goals are twofold: (i) to provide reproducible reference results that span the multilingual, code-switched, and mixed–emotion properties of the corpus introduced in Sections I-III, and (ii) to clarify how speaker factors (gender, age class) and personality traits (MBTI subset) modulate LDL performance. 

\subsubsection{Evaluation Protocol}
The EM$^2$LDL corpus was partitioned into training, validation, and test sets in a speaker-independent manner to ensure robust generalization. All experiments were conducted at the \emph{utterance level} with ground-truth emotion \emph{label distributions}. We designed four complementary evaluation settings to probe corpus factors:

\begin{itemize}
\item \textbf{Experiment 1: Overall Baseline Performance on the EM$^{2}$LDL.} A speaker-independent split of the full corpus into training/validation/test partitions at a 70/10/20 ratio (7:1:2) by speaker ID. No speaker overlap occurs across partitions.
\item \textbf{Experiment 2: Gender-Effect Investigation on the EM$^{2}$LDL.} Cross-validation (CV) was performed by training on male speakers (1,022 utterances) and testing on female speakers (2,976 utterances), and vice versa, to examine gender-specific emotional expression patterns.
\item \textbf{Experiment 3: Age-Effect Investigation on the EM$^{2}$LDL.} CV was conducted by training on youth speakers (3,220 utterances) and testing on non-youth speakers (778 utterances), and vice versa, to investigate age-related differences in emotional speech.
\item \textbf{Experiment 4: Personality-Effect Investigation on the EM$^{2}$LDL.} A leave-one-out personality (LOPO) CV was performed on a subset of speakers with MBTI annotations, covering four personality types (IN, IS, EN, ES). Each fold trained on three personality types and tested on the held-out type, resulting in a 4-fold CV.
\end{itemize}

\subsubsection{Acoustic Features}
To capture robust speech representations, we utilized self-supervised learning (SSL) models as front-end feature extractors, selected for their ability to model complex acoustic patterns in multilingual and code-switched speech. The following SSL models were used: Hubert-base-CN\footnotemark[2], Hubert-large-CN\footnotemark[3], wav2vec-base-CN\footnotemark[4], wav2vec-large-CN\footnotemark[5], WavLM-base-EN\footnotemark[6], WavLM-large-EN\footnotemark[7], Hubert-base-EN\footnotemark[8], Hubert-large-EN\footnotemark[9], Whisper-Base\footnotemark[10], and Whisper-Large-V3\footnotemark[11]. For each utterance, SSL embeddings were extracted using the respective model’s feature extraction pipeline, yielding 768-dimensional (base models) or 1024-dimensional (large models) representations. These embeddings were averaged across the temporal dimension to produce one fixed-length feature vector per utterance, which served as input to the downstream model.
\footnotetext[2]{\url{https://huggingface.co/TencentGameMate/chinese-hubert-base}}
\footnotetext[3]{\url{https://huggingface.co/TencentGameMate/chinese-hubert-large}}
\footnotetext[4]{\url{https://huggingface.co/TencentGameMate/chinese-wav2vec2-base}}
\footnotetext[5]{\url{https://huggingface.co/TencentGameMate/chinese-wav2vec2-large}}
\footnotetext[6]{\url{https://huggingface.co/microsoft/wavlm-base}}
\footnotetext[7]{\url{https://huggingface.co/microsoft/wavlm-large}}
\footnotetext[8]{\url{https://huggingface.co/facebook/hubert-base-ls960}}
\footnotetext[9]{\url{https://huggingface.co/facebook/hubert-large-ls960-ft}}
\footnotetext[10]{\url{https://huggingface.co/openai/whisper-base}}
\footnotetext[11]{\url{https://huggingface.co/openai/whisper-large-v3}}

\subsubsection{Model Architecture and Training}
Our baseline model adopts a minimalist architecture to ensure that performance differences primarily reflect the choice of front-end representations. For each utterance, frame-level features are extracted using a pre-trained speech SSL model. These features are mean-pooled along the temporal dimension to obtain a fixed-dimensional utterance embedding $\mathbf{h} \in \mathbb{R}^{768}$. This embedding is projected via a single linear layer to a 32-dimensional output, with log-softmax applied to yield predicted log-probabilities $\hat{\mathbf{z}} \in \mathbb{R}^{32}$:

\begin{equation}
\hat{\mathbf{z}} = \log \mathrm{softmax}(\mathbf{W}\mathbf{h} + \mathbf{b}).
\end{equation}

Model training jointly minimizes two objectives: (i) the Kullback–Leibler (KL) divergence between the predicted distribution $\hat{\mathbf{p}} = \exp(\hat{\mathbf{z}})$ and the target label distribution $\mathbf{y}$ (Sec.~III-C), and (ii) an auxiliary cosine similarity loss weighted by $\lambda_{\text{cos}}$ (default $0.1$) to encourage angular alignment between the predicted and target distributions:
\begin{align}
\mathcal{L}_{\text{KL}} &= \sum_{c=1}^{32} y_c \log \frac{y_c}{\hat{p}_c + \epsilon}, \\
\mathcal{L}_{\text{cos}} &= 1 - \frac{\mathbf{y}\cdot\hat{\mathbf{p}}}{\lVert \mathbf{y} \rVert_2 \lVert \hat{\mathbf{p}} \rVert_2 + \epsilon}, \\
\mathcal{L} &= \mathcal{L}_{\text{KL}} + \lambda_{\text{cos}}\mathcal{L}_{\text{cos}}.
\end{align}

The model is trained using the Adam optimizer with a learning rate of 0.0001 and batch size of 16. The maximum number of epochs is set to 50, and early stopping is implemented with a patience of 5 based on validation loss. Training and evaluation are conducted on a single NVIDIA GPU with mixed-precision acceleration. Audio augmentation is deliberately omitted in all baseline experiments to ensure that corpus factors, rather than regularization effects, dominate the observed differences. For model selection, the best checkpoint on the validation set is determined by the lowest $\mathcal{L}_{\text{KL}}$; in case of ties, the model with higher cosine similarity is selected. Early stopping is applied on the basis of this criterion, and the selected checkpoint is evaluated on the test set.

\subsubsection{Evaluation Metrics}
To comprehensively assess model performance, we report a panel of distributional distances and similarities commonly used in LDL evaluation \cite{khelifa2025label,geng2016label}. 

Let $\mathbf{y}=(y_1,\ldots,y_C)$ denote the reference distribution and $\hat{\mathbf{p}}=(\hat{p}_1,\ldots,\hat{p}_C)$ the system prediction over $C{ = }32$ emotions (both nonnegative and summing to~1). A small constant $\epsilon$ (e.g., $10^{-8}$) is added where needed to avoid division by~0.

\begin{align}
\text{Chebyshev}~(\downarrow) &= \max_{c} |y_c - \hat{p}_c| \label{eq:cheb}\\[2pt]
\text{Clark}~(\downarrow) &= \sqrt{\sum_{c} \left(\frac{y_c - \hat{p}_c}{y_c + \hat{p}_c + \epsilon}\right)^2} \label{eq:clark}\\[2pt]
\text{Canberra}~(\downarrow) &= \sum_{c} \frac{|y_c - \hat{p}_c|}{|y_c| + |\hat{p}_c| + \epsilon} \label{eq:canb}\\[2pt]
\text{Kullback--Leibler}~(\downarrow) &= \sum_{c} y_c \log \frac{y_c}{\hat{p}_c + \epsilon} \label{eq:kl}\\[2pt]
\text{Cosine}~(\uparrow) &= \frac{\mathbf{y}\cdot\hat{\mathbf{p}}}{\lVert \mathbf{y} \rVert_2 \lVert \hat{\mathbf{p}} \rVert_2 + \epsilon} \label{eq:cos}\\[2pt]
\text{Intersection}~(\uparrow) &= \sum_{c} \min(y_c, \hat{p}_c) \label{eq:inter}
\end{align}

\textcolor{black}{These metrics evaluate different aspects of the alignment between predicted and reference emotion distributions: }
\begin{itemize}
\item \textbf{Chebyshev} measures the maximum pointwise deviation, highlighting the largest error in any single emotion;
\item \textbf{Clark} assesses the normalized squared differences, emphasizing relative errors across the distribution; 
\item \textbf{Canberra} quantifies the sum of normalized absolute differences, sensitive to small values; 
\item \textbf{KL} evaluates the information-theoretic divergence, capturing how much one distribution differs from another in terms of entropy; 
\item \textbf{Cosine} measures the angular similarity, focusing on the directional agreement between vectors;  
\item \textbf{Intersection} quantifies the overlapping mass, indicating the shared probability between distributions. 
\end{itemize}

Collectively, they provide a robust evaluation framework that captures both divergence and similarity aspects of the predicted emotion label distributions. The results of these experiments (reported in subsequent subsections) demonstrate the effectiveness of the EM$^2$LDL corpus and the LDL framework in modeling complex, multilingual, and mixed-emotion aspects.

\begin{table*}[t]
\centering
\caption{\textcolor{black}{Performance of LDL Across SSL Front-Ends on the EM$^2$LDL Corpus (Speaker-Independent Split). $\downarrow$: lower is better, $\uparrow$: higher is better. $^{\dagger}$: significantly worse than the best model for that metric (paired Wilcoxon; Holm-adjusted $p<0.05$).}}
\label{tab:ssl_baseline}
\small
\resizebox{\textwidth}{!}{
\begin{tabular}{lcccccc}
\toprule
\textbf{SSL} & \textbf{Cheby. ($\downarrow$)} & \textbf{Clark ($\downarrow$)} & \textbf{Can. ($\downarrow$)} & \textbf{KL ($\downarrow$)} & \textbf{Cos. ($\uparrow$)} & \textbf{Int. ($\uparrow$)} \\
\midrule
HuBERT-base-CN & 0.1582 & 4.9673$^{\dagger}$ & 26.3591$^{\dagger}$ & 0.8492$^{\dagger}$ & 0.7951$^{\dagger}$ & 0.5212$^{\dagger}$ \\
HuBERT-large-CN & 0.1576 & 4.9770$^{\dagger}$ & 26.4294$^{\dagger}$ & 0.8535$^{\dagger}$ & 0.7933$^{\dagger}$ & 0.5261$^{\dagger}$ \\
Wav2vec2-base-CN & 0.1576$^{\dagger}$ & 4.9723$^{\dagger}$ & 26.3977$^{\dagger}$ & 0.8478$^{\dagger}$ & 0.7948$^{\dagger}$ & 0.5248$^{\dagger}$ \\
Wav2vec2-large-CN & 0.1568 & 4.9812$^{\dagger}$ & 26.4667$^{\dagger}$ & 0.8481$^{\dagger}$ & 0.7940$^{\dagger}$ & 0.5303$^{\dagger}$ \\
WavLM-base-EN & 0.1580 & 4.9680$^{\dagger}$ & 26.3500$^{\dagger}$ & 0.8523$^{\dagger}$ & 0.7948$^{\dagger}$ & 0.5221$^{\dagger}$ \\
WavLM-large-EN & 0.1568 & 4.9845$^{\dagger}$ & 26.4875$^{\dagger}$ & 0.8480$^{\dagger}$ & 0.7942$^{\dagger}$ & 0.5324$^{\dagger}$ \\
HuBERT-base-EN & 0.1574$^{\dagger}$ & 4.9877$^{\dagger}$ & 26.5224$^{\dagger}$ & 0.8501$^{\dagger}$ & 0.7926$^{\dagger}$ & \textbf{0.5331} \\
HuBERT-large-EN & \textbf{0.1565} & 4.9784$^{\dagger}$ & 26.4362$^{\dagger}$ & \textbf{0.8462} & \textbf{0.7955} & 0.5306$^{\dagger}$ \\
Whisper-Base & 0.1618$^{\dagger}$ & \textbf{4.9612} & \textbf{26.2996} & 0.8564$^{\dagger}$ & 0.7938$^{\dagger}$ & 0.5150$^{\dagger}$ \\
Whisper-Large-v3 & 0.1584 & 4.9688$^{\dagger}$ & 26.3529$^{\dagger}$ & 0.8491$^{\dagger}$ & 0.7949$^{\dagger}$ & 0.5235$^{\dagger}$ \\\bottomrule
\end{tabular}
}
\end{table*}

\subsection{Overall Baseline Performance on the EM$^{2}$LDL}
To evaluate the baselines of the EM$^2$LDL corpus for mixed-emotion recognition, we conducted a comprehensive analysis using multiple SSL front-ends. The performance of these SSLs was assessed on a speaker-independent split of the corpus. Table~\ref{tab:ssl_baseline} presents the results across six metrics: Chebyshev (Cheby.), Clark, Canberra (Can.), Kullback-Leibler (KL), Cosine similarity (Cos.), and Intersection (Int.). 

\textcolor{black}{In addition, we performed paired significance tests on the utterance-level scores exported for each metric. For each non-best model, we conducted a paired Wilcoxon signed-rank test against the best model over the same set of test utterances, thereby controlling for utterance-specific variability. We utilized one-sided alternatives to test whether the non-best model was worse than the best.} \textcolor{black}{To control the family-wise error rate across the nine pairwise comparisons for each metric, we applied Holm–Bonferroni correction ($\alpha $ = 0.05).} \textcolor{black}{In Table~\ref{tab:ssl_baseline}, the best result in each column is shown in boldface, and a superscript $^{\dagger}$ is added to entries that are significantly worse than the best after multiple-comparison correction. Entries without superscripts are not significantly different from the best under the adopted test.}

The results demonstrate that HuBERT-large-EN achieves the best performance across three metrics: Chebyshev (0.1565), K-L (0.8462), and Cosine similarity (0.7955). This suggests that HuBERT-large-EN effectively captures the nuanced emotion distributions in the EM$^2$LDL corpus, likely due to its robust pre-training on English speech data, which aligns with a significant portion of the corpus. Notably, HuBERT-base-EN outperforms other models in the Intersection metric (0.5331), indicating strong agreement with ground-truth emotion distributions. Conversely, Whisper-Base yields the best results for Clark (4.9612) and Canberra (26.2996) distances, suggesting its capability to minimize cumulative differences in emotion distributions, possibly due to its broad pre-training across diverse audio data.

Among the Chinese-specific models, Wav2vec2-large-CN and HuBERT-large-CN exhibit competitive performance, with Wav2vec2-large-CN achieving a Chebyshev distance of 0.1568 and HuBERT-large-CN scoring 0.5261 in Intersection. These results highlight the models' ability to handle tonal languages like Mandarin and Cantonese, though they slightly underperform compared to English-focused models, likely due to the linguistic complexity introduced by code-switching. WavLM-large-EN also shows strong performance (0.5324 in Intersection), reinforcing the advantage of large-scale SSL models in capturing emotional nuances across languages.

Unexpectedly, Whisper-Large-v3 does not outperform smaller models like HuBERT-base-EN or Wav2vec2-large-CN in most metrics, despite its larger architecture. This may indicate that the model's broad pre-training across multiple tasks dilutes its specialization for fine-grained emotion distribution modeling in multilingual contexts. Additionally, the marginal performance differences among models (e.g., Chebyshev ranging from 0.1565 to 0.1618) suggest that the EM$^2$LDL corpus poses a challenging benchmark, where even advanced SSL models struggle to fully capture the complexity of mixed emotions and intra-utterance code-switching.

These findings align with prior research on SSL models for speech processing, where models like HuBERT and Wav2vec2 have demonstrated strong generalization across diverse tasks \cite{wang2021fine}. However, the EM$^2$LDL corpus's unique combination of multilingualism and mixed-emotion annotations reveals limitations in current SSL approaches, particularly in handling code-switched utterances. The results underscore the need for models tailored to dynamic linguistic environments, as discussed in Section II. Furthermore, the consistently high performance of English-focused models may reflect the corpus's balanced representation of English alongside Mandarin and Cantonese, though further analysis is needed to explore potential biases in language distribution.

\begin{table*}[!t]
\centering
\caption{Gender-Cross Evaluation Performance of SSL Front-Ends on EM$^2$LDL. Each model is evaluated under Male (M)$\rightarrow$Female (F) (Training$\rightarrow$Test) and F$\rightarrow$M settings. $\downarrow$: lower is better, $\uparrow$: higher is better.}
\label{tab:gender_cross}
\resizebox{\textwidth}{!}{
\begin{tabular}{lcccccc|cccccc}
\toprule
\multirow{2}{*}{\textbf{SSL}} & \multicolumn{6}{c|}{\textbf{M$\rightarrow$F}} & \multicolumn{6}{c}{\textbf{F$\rightarrow$M}} \\
& \textbf{Cheby. ($\downarrow$)} & \textbf{Clark ($\downarrow$)} & \textbf{Can. ($\downarrow$)} & \textbf{KL ($\downarrow$)} & \textbf{Cos. ($\uparrow$)} & \textbf{Int. ($\uparrow$)} & \textbf{Cheby. ($\downarrow$)} & \textbf{Clark ($\downarrow$)} & \textbf{Can. ($\downarrow$)} & \textbf{KL ($\downarrow$)} & \textbf{Cos. ($\uparrow$)} & \textbf{Int. ($\uparrow$)} \\

\midrule
HuBERT-base-CN       & 0.1606 & 5.0047 & 26.6912 & 0.8792 & 0.7941 & 0.5139 & 0.1637 & 4.9926 & 26.6426 & 0.9032 & 0.7585 & 0.5049 \\
HuBERT-large-CN      & 0.1599 & 5.0103 & 26.7643 & 0.8800 & 0.7933 & 0.5154 & 0.1650 & 4.9816 & 26.5676 & 0.9050 & 0.7583 & 0.4976 \\
Wav2vec2-base-CN     & 0.1794 & 4.9673 & 26.4428 & 0.9171 & 0.7852 & 0.4673 & 0.1635 & 4.9987 & 26.7177 & 0.9080 & 0.7565 & 0.5034 \\
Wav2vec2-large-CN    & 0.1603 & 5.0126 & 26.7563 & 0.8831 & 0.7926 & 0.5164 & 0.1631 & 4.9852 & 26.5654 & 0.9026 & 0.7603 & 0.5017 \\
WavLM-base-EN        & 0.1608 & 5.0054 & 26.7338 & 0.8827 & 0.7931 & 0.5110 & 0.1629 & 4.9845 & 26.5731 & 0.9036 & 0.7592 & 0.5013 \\
WavLM-large-EN       & 0.1603 & 5.0085 & 26.7349 & 0.8798 & 0.7935 & 0.5151 & 0.1627 & 5.0008 & 26.6977 & 0.9054 & 0.7583 & 0.5090 \\
HuBERT-base-EN       & 0.1599 & 5.0104 & 26.7416 & 0.8796 & 0.7936 & 0.5170 & 0.1671 & 5.0029 & 26.7362 & 0.9101 & 0.7525 & 0.5059 \\
HuBERT-large-EN      & 0.1597 & 5.0185 & 26.8301 & 0.8835 & 0.7922 & 0.5183 & 0.1628 & 4.9799 & 26.5285 & 0.8990 & 0.7612 & 0.5003 \\
Whisper-Base         & 0.2270 & 4.9563 & 26.2795 & 1.0454 & 0.6936 & 0.4067 & 0.1648 & 4.9876 & 26.6240 & 0.9094 & 0.7576 & 0.4975 \\
Whisper-Large-V3     & 0.1797 & 4.9748 & 26.4526 & 0.9078 & 0.7812 & 0.4769 & 0.1631 & 4.9965 & 26.6817 & 0.9064 & 0.7577 & 0.5045 \\
\bottomrule
\end{tabular}
}
\end{table*}

\begin{figure*}[t]  
    \centering
    \includegraphics[width=\textwidth]{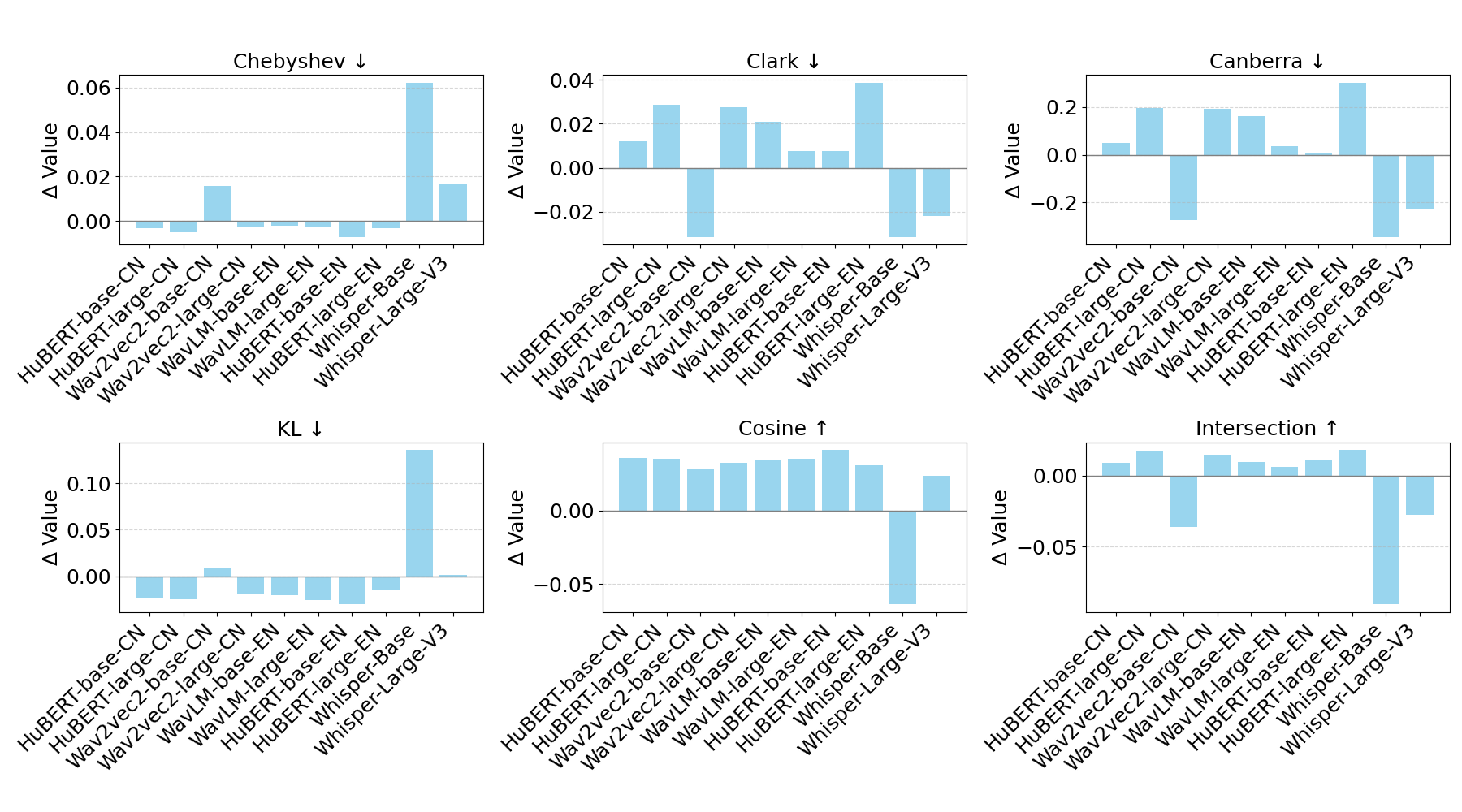}  
    \caption{Gender cross-validation performance differences ($\Delta $Value: F tested $-$ M tested) across label distribution metrics.}
    \label{fig:gendercross}
\end{figure*}

\subsection{Gender-Effect Investigation on the EM$^{2}$LDL}
To investigate model robustness across speaker demographics, we conducted a gender CV evaluation using the EM$^2$LDL corpus. In this setup, each SSL front-end was trained exclusively on one gender (either male or female) and tested on the opposite gender. This configuration allows us to assess the generalization capacity of models when exposed to gender-specific variations in emotional speech expression.

Table~\ref{tab:gender_cross} lists the LDL performances under gender cross-validation settings. To further interpret these results, Fig.~\ref{fig:gendercross} summarizes the tested performance difference ($\Delta$ = Female $-$ Male) across six evaluation metrics. Notably, these metrics follow different optimization directions: for Chebyshev, Clark, Canberra, and KL divergence, lower values are desirable ($\downarrow$), while for Cosine and Intersection, higher values indicate better alignment with ground truth ($\uparrow$). Accordingly, a negative $\Delta$ indicates improvement on female speech for divergence metrics, and a positive $\Delta$ indicates improvement on female speech for similarity metrics.

As visualized in Fig.~10, several models demonstrated consistent gains on female test data. HuBERT-base-EN, for example, achieves the most favorable delta across metrics, with improvements on all four divergence measures (e.g., $\Delta$KL = $-0.0305$, $\Delta$Cheby. = $-0.0072$) and gains in both similarity metrics ($\Delta$Cos. = $+0.0411$, $\Delta$Int. = $+0.0111$). These results suggest that the model better captures emotion distributions when tested on female utterances, potentially due to higher expressivity or acoustic clarity in female vocal patterns. In contrast, Whisper-Base exhibits substantial degradation when evaluated on female data, with a large positive delta in KL ($\Delta$KL = $+0.1360$) and negative shifts in Cosine ($-0.0640$) and Intersection ($-0.0908$), indicating its limited adaptability to gender variance despite multilingual pretraining. \textcolor{black}{This can potentially be attributed to two factors. First, Whisper is pre-trained on ASR, and in our minimalist pipeline, we mean-pool frame-level embeddings and utilize a single linear head, which may dilute the fine-grained prosody needed for label-distribution emotion modeling. Second, model capacity affects demographic transfer: base-sized encoders suffer larger drops than large counterparts in gender shifts.}

Interestingly, these observations support three key findings. First, female speech often yields more precise emotion distribution predictions under multiple SSLs. Models such as HuBERT-large-EN, HuBERT-base-EN, and WavLM variants show consistent improvements when tested on female speakers, as reflected by negative deltas in divergence metrics and positive gains in similarity metrics. This suggests that emotional prosody in female speech is often characterized by greater pitch variation and more dynamic intonation, which may facilitate finer alignment with human-annotated label distributions.

Second, gender-induced performance gaps are model-specific and metric-sensitive. While Wav2vec2-base-CN, Wav2vec2-large-CN, and Whisper models reveal mixed or adverse trends on female speech, HuBERT and WavLM models exhibit stronger gender robustness. Moreover, the extent of variation is not uniform across metrics: Whisper-Base shows the largest divergence penalty ($\Delta$KL = $+0.1360$) and yet a relatively minor change in Clark distance ($-0.0313$), underscoring the need for multi-metric interpretation.

Third, the pretraining strategy plays a critical role in gender generalization. English-centric HuBERT models outperform Chinese-pretrained SSL variants, even under cross-lingual and code-switched conditions. This implies that pretraining on richly emotional and prosodically diverse speech corpora may endow models with improved generalizability to gendered variations in emotional expression.

Overall, despite the probabilistic and soft-label nature of LDL, speaker gender remains a salient factor influencing model behavior. While the EM$^2$LDL corpus mitigates bias through balanced gender representation and diversified emotional labeling, SSL front-ends differ substantially in how well they generalize across gender domains. This serves as a motivation for future work on gender-aware adaptation, demographic-invariant training objectives, and cross-gender normalization techniques to further enhance equity and robustness in multilingual affective computing systems.

\begin{figure*}[t]  
\centering
\includegraphics[width=\textwidth]{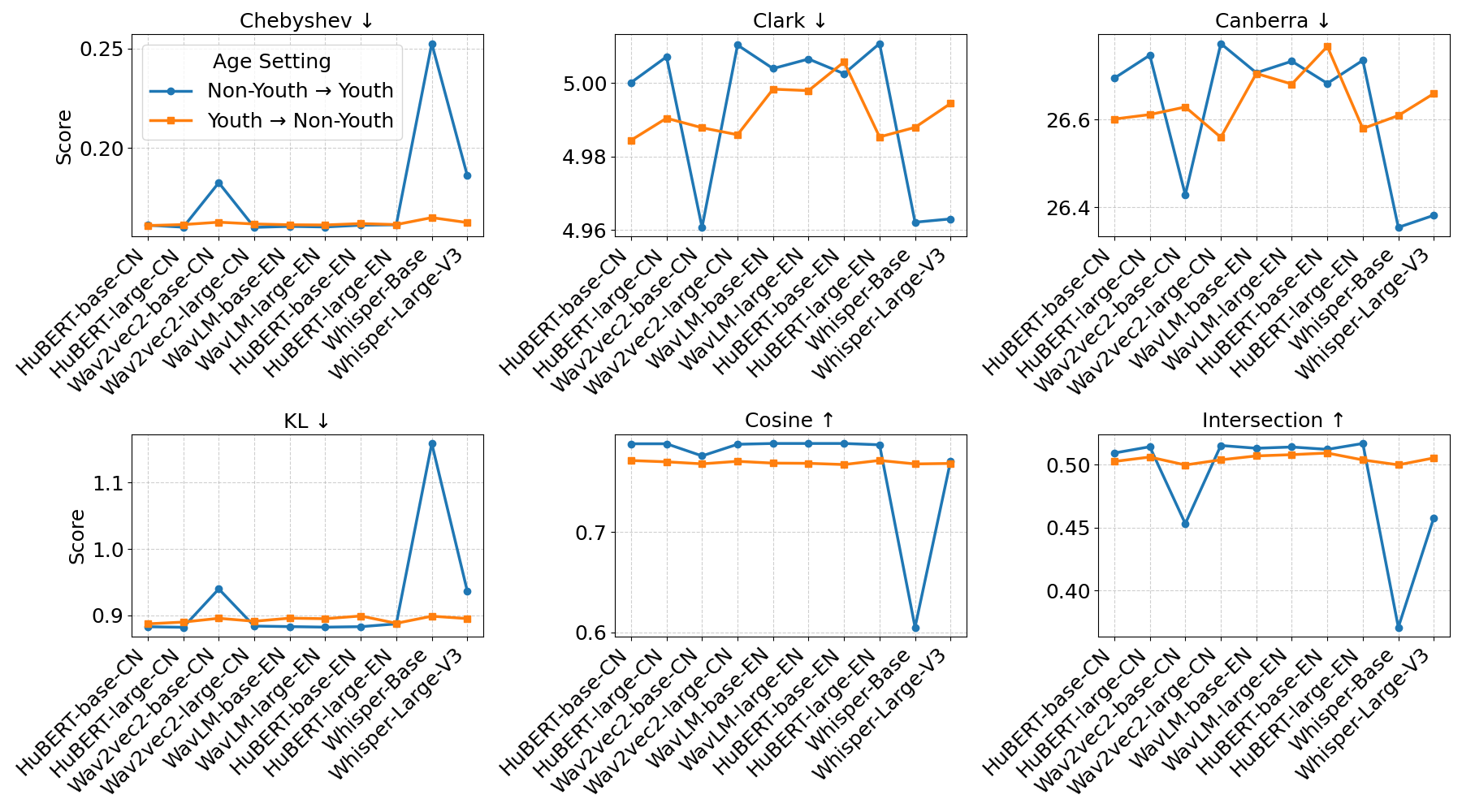}  
\caption{Age cross-validation performance of SSL models on EM$^2$LDL. Each model is evaluated under Non-Youth$\rightarrow$Youth (Training$\rightarrow$Test) and Youth$\rightarrow$Non-Youth settings. $\downarrow$: lower is better, $\uparrow$: higher is better.}
\label{fig:agecross}
\end{figure*}

\subsection{Age-Effect Investigation on the EM$^{2}$LDL}
To further explore how speaker age affects multilingual mixed-emotion recognition, we conducted age CV experiments using two non-overlapping groups: \textit{Youth} and \textit{Non-Youth}. Specifically, we trained each SSL model on one group and evaluated its generalization on the other, and vice versa. Evaluation was performed using six label distribution-based metrics, as summarized and illustrated in Fig.~\ref{fig:agecross}.

Compared to the gender-based setting in Section IV-C, the age effect manifests more subtly yet remains informative. Overall, models trained on non-youth data and tested on youth (\textit{Non-Youth$\rightarrow$Youth}) tend to exhibit greater performance degradation than in the reverse direction. For example, Whisper-Base shows a KL divergence of 1.1588 when trained on non-youth speakers and evaluated on youth, compared to 0.9643 when trained on youth and tested on non-youth. Similar asymmetries are consistently observed across models and metrics, particularly for distributional distance measures such as Chebyshev and Canberra.

These results suggest that emotional speech distributions expressed by youth speakers are relatively more variable or less well-represented in models trained solely on older demographics. In contrast, when trained on youth data, models appear to generalize better to the non-youth test set, possibly because youth vocal expressions cover a broader affective range or exhibit greater acoustic variability. This directional asymmetry reveals that youth-derived emotional expressions may inherently pose more challenges for cross-age generalization.

Furthermore, model capacity plays a key role in mediating age-induced performance drops. Large models such as HuBERT-large-EN and WavLM-large-EN maintain a relatively stable performance in both directions, with minimal fluctuation in KL divergence and cosine similarity. In contrast, Whisper-Base and Wav2vec2-base-CN demonstrate more pronounced degradation, reaffirming the advantage of pretraining scale and diversity in tackling age-related distribution shifts.

Interestingly, while the similarity-based metrics (Cosine, Intersection) remain relatively stable, distribution-sensitive metrics (e.g., KL, Canberra) are more affected under age mismatch. This implies that although the overall emotion space may remain comparable, the fine-grained label distributions are more susceptible to variation across age. In practical terms, this suggests that systems deployed in youth-oriented settings may benefit from explicit age modeling or adaptive regularization to bridge these latent distribution gaps.

In summary, although the age effect is not as dramatic as gender-induced variations, it reveals key asymmetries in emotional generalization. The results emphasize the importance of including age-diverse data during training and call for future research into age-conditioned modeling and data augmentation strategies for more robust affective computing systems.

\begin{figure*}[t]  
\centering
\includegraphics[width=\textwidth]{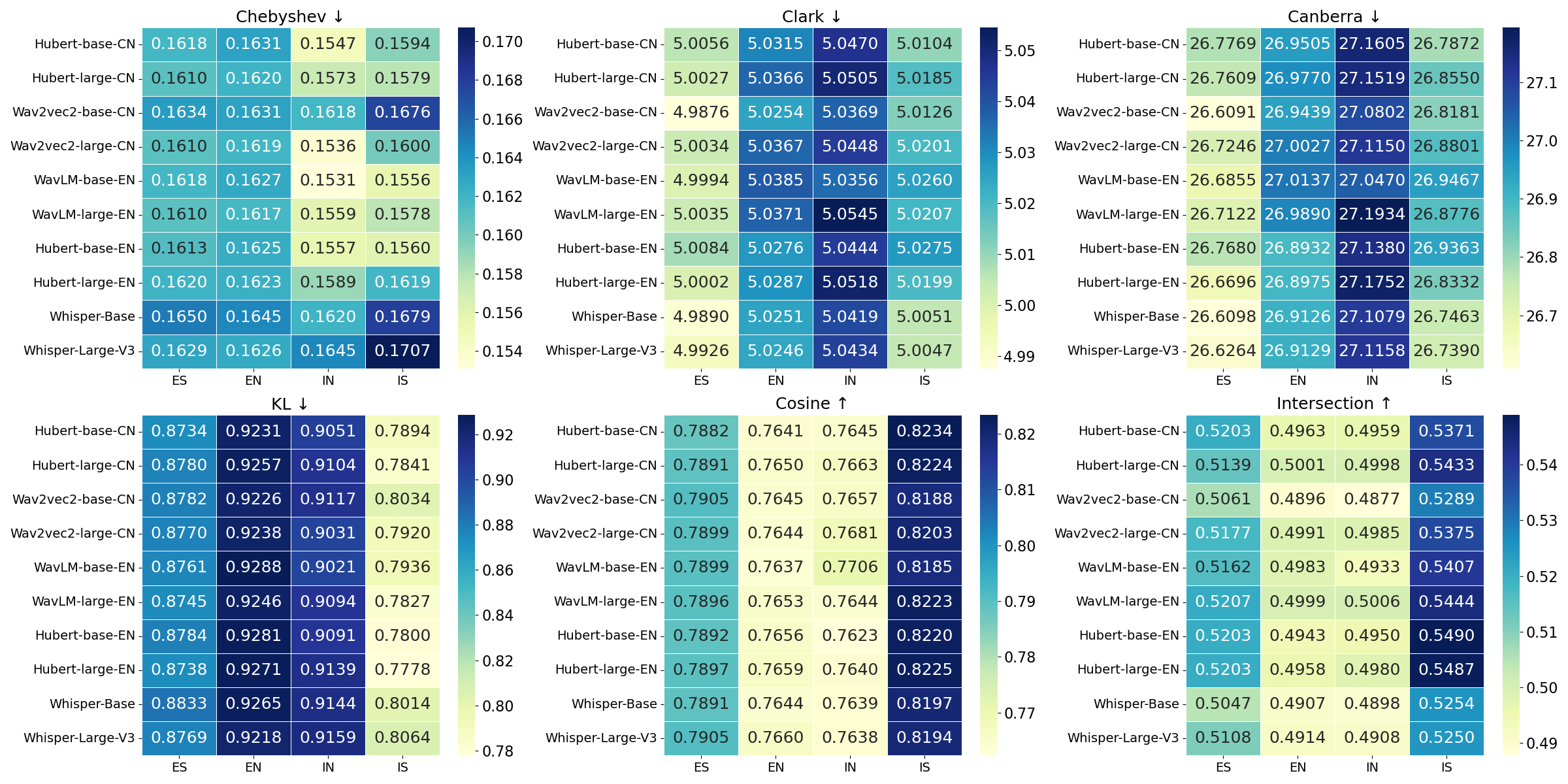}  
\caption{Personality-cross validation performance of SSL models on EM$^2$LDL. Each model is tested by one MBTI subset and trained on the others. $\downarrow$: lower is better, $\uparrow$: higher is better.}
\label{fig:mbticross}
\end{figure*}

\subsection{Personality-Effect Investigation on the EM$^{2}$LDL}
To investigate how individual personality traits impact emotion recognition under the label distribution learning paradigm, we conducted a cross-personality evaluation based on the MBTI framework. Specifically, we selected four representative personality types—ES, EN, IS, and IN—and performed a leave-one-type-out experiment. In each split, samples corresponding to one MBTI type were reserved exclusively for testing, while the remaining three types constituted the training set. This setup allows us to analyze how well models generalize to unseen emotional expression styles associated with different personality dimensions, particularly the extraversion–introversion (E/I) and sensing–intuition (S/N) axes.

Figure~\ref{fig:mbticross} shows the performance of ten SSL front-ends across the four personality-specific test sets. The evaluation metrics for each test personality are reported. Brighter cells in the Chebyshev, Clark, Canberra, and KL subplots indicate lower distribution divergence (better performance), while darker cells in the Cosine and Intersection subplots represent higher alignment between predicted and ground-truth distributions.

Several intriguing insights arise from our personality-aware evaluation. Notably, the IS (Introversion + Sensing) speakers demonstrate the most robust generalization across all evaluation metrics. When averaged over the ten SSL models, IS-type samples yield the lowest KL divergence (0.7911) and Chebyshev distance (0.1626), while also attaining the highest Cosine similarity (0.8209) and Intersection score (0.5380). Such consistency suggests that IS speakers’ emotional expressions—grounded in concrete sensory perception and inward-focused processing—may manifest in more stable and predictable acoustic patterns. These attributes likely enable emotion predictors to better align their output distributions with the true emotion landscape.

In contrast, models encounter substantial difficulty when generalizing to speakers with intuitive personality traits. Both EN (Extraversion + Intuition) and IN (Introversion + Intuition) types present challenging test scenarios. EN speakers, marked by expressiveness and abstract ideation, exhibit the highest average KL divergence (0.9252) and Chebyshev distance (0.1626), coupled with the lowest similarity scores. Similarly, IN speakers—though slightly better on divergence—display low consistency in Cosine (0.7644) and Intersection (0.4959), indicating that their emotionally nuanced, introspective speech may elude straightforward modeling. These findings align with psychological theories suggesting that intuitive individuals often rely on abstract cues and less behaviorally anchored expressions, which pose challenges for data-driven predictors.

Interestingly, ES (Extraversion + Sensing) speakers occupy a middle ground. While extraverted and socially expressive, their reliance on present-moment sensory information may yield emotionally vivid but structurally regular cues. This is reflected in their near-average divergence and similarity metrics across the board, suggesting that models neither struggle nor excel with this group—likely due to the balance between external expressiveness and perceptual grounding.

These results demonstrate that personality-linked affective styles subtly but systematically shape the output distribution quality in SER tasks. Importantly, such variation persists even under controlled lexical and acoustic conditions, underscoring the nontrivial role of speaker personality in emotion modeling. Moving forward, incorporating soft personality conditioning—whether through trait-aware embeddings or auxiliary learning objectives—may offer a promising pathway toward more resilient and personalized affective computing systems.

\section{Future Directions}
The development of EM$^2$LDL highlights both opportunities and challenges in multilingual mixed-emotion recognition under the LDL paradigm. By introducing emotion distributions instead of discrete labels, and by covering multiple languages, dialects, and speaker traits, EM$^2$LDL encourages a shift toward more ecologically valid affective computing. Nevertheless, this work also reveals several limitations that suggest important directions for future research.

One notable limitation concerns the modeling of code-switching behavior. Although EM$^2$LDL contains a substantial portion of bilingual utterances, particularly co-switched mixtures, our current experiments do not explicitly model language-switching dynamics. Code-switching introduces shifts in acoustic, phonetic, and prosodic structure, which can influence how emotions are expressed and perceived \cite{williams2020bilinguals,olson2016impact}. As such, future models may benefit from integrating code-switch-aware front-ends or multilingual encoders capable of handling dynamic linguistic transitions.

Beyond linguistic variation, the perceptual nature of our annotation process introduces another underexplored dimension: rater personality. Since EM$^2$LDL provides personality trait labels for human annotators, this opens up the possibility of incorporating rater disposition into future modeling. Prior work has shown that personality traits such as extraversion and neuroticism affect how emotional expressions are interpreted \cite{bono2007personality,costa1980influence}. Leveraging this information could enable the design of rater-aware learning objectives or label calibration schemes that better reflect interrater subjectivity.

Another concern relates to demographic imbalance. Although speaker diversity was an explicit design goal, the final corpus still exhibits skewed distributions across gender and age groups. Such imbalance can lead to model bias and reduced generalizability. To mitigate this, future extensions may explore data resampling strategies, adversarial domain balancing, or fairness-aware training frameworks.

In addition, the current study focuses solely on acoustic cues. While this decision enables focused analysis of paralinguistic features, it limits the ability to model linguistic-semantic interactions—particularly in emotionally blended scenarios where voice and content may convey conflicting or complementary affects. A promising direction would be to incorporate automatic speech recognition outputs and explore multimodal emotion recognition systems that align acoustic tone with semantic meaning \cite{sebe2005multimodal,abdullah2021multimodal}.

These technical directions also invite broader theoretical inquiry. For example, the structure of EM$^2$LDL supports analysis of how speaker-intrinsic factors such as personality, age, and gender influence emotional expression and recognition. This could facilitate the development of personalized SER systems that adapt to the expressive styles of individual users, as well as enable empirical testing of socio-cognitive theories of affect and interpersonal variability.

In short, the EM$^2$LDL corpus is not merely a resource but rather an important step toward a richer modeling paradigm—one that recognizes emotion as a dynamic, multilingual, socially contextualized phenomenon. Building on this foundation will require the joint efforts of computational, linguistic, and psychological perspectives to design affective systems that are both inclusive and adaptive across human diversity.

\section{Conclusion}
This study introduced EM$^{2}$LDL, a novel multilingual speech corpus designed to advance SER by addressing the complexities of mixed emotions and intra-utterance code-switching in multilingual contexts. By integrating English, Mandarin, and Cantonese, and incorporating spontaneous code-switched utterances from sociolinguistic settings such as Hong Kong and Macao, EM$^{2}$LDL directly addresses the gap left by predominantly monolingual and single-label SER corpora. \textcolor{black}{Our evaluation results substantiate this contribution: baseline experiments show that even state-of-the-art SSL models struggle to capture the fine-grained distributions in EM$^{2}$LDL, highlighting the need for corpora that challenge models with linguistically dynamic and emotionally complex content. Moreover, gender and age cross-validations reveal systematic asymmetries, such as youth-to-non-youth transfer being more robust than the reverse, while personality-based evaluations confirm that label distributions capture individual variability beyond single-label annotations. These findings collectively demonstrate that EM$^{2}$LDL provides measurable benefits for advancing robust and ecologically valid SER research.}

The significance of EM$^{2}$LDL also lies in its ability to support the development of SER models that generalize across linguistic, cultural, and demographic boundaries. The corpus encompasses 231 unique speakers with relatively balanced gender and age distributions, and includes MBTI personality traits for a subset of speakers. \textcolor{black}{This diversity strengthens its applicability for studying the interplay between speaker characteristics and emotional expression, positioning EM$^{2}$LDL as both a benchmark resource and a catalyst for developing more inclusive and psychologically grounded affective computing systems.}

\textcolor{black}{EM$^{2}$LDL also points to promising directions for future work. Its focus on intra-utterance code-switching underscores the need for models that explicitly account for dynamic linguistic transitions, potentially through code-switch-aware front-ends or multilingual encoders. The availability of rater personality data offers opportunities for rater-aware objectives to mitigate subjectivity in annotations. Finally, demographic imbalances highlight the importance of fairness-aware modeling to ensure equitable representation across user groups.}

\textcolor{black}{In conclusion, EM$^{2}$LDL represents a significant step forward in addressing the limitations of traditional SER corpora. By offering a comprehensive dataset that captures the complexity of emotional and linguistic dynamics in multilingual settings, and by validating these contributions with extensive experimental evidence, this work not only advances the theoretical understanding of mixed emotions but also provides practical implications for building empathetic, context-aware technologies. Collectively, these contributions establish EM$^{2}$LDL as the first resource to systematically integrate multilingualism, intra-utterance code-switching, and mixed-emotion annotation under the LDL paradigm, paving the way for SER systems that are more inclusive, adaptive, and ecologically valid.}

\bibliographystyle{IEEEtran}
\bibliography{refs}
\end{document}